\documentclass{article}

\usepackage{PRIMEarxiv}

\usepackage[utf8]{inputenc} 
\usepackage[T1]{fontenc}    
\usepackage{float}
\usepackage{hyperref}       
\usepackage{url}            
\usepackage{booktabs}       
\usepackage{amsfonts}       
\usepackage{nicefrac}       
\usepackage{microtype}      
\usepackage{lipsum}
\usepackage{fancyhdr}       
\usepackage{graphicx}       
\usepackage{amsmath}
\graphicspath{{media/}}     
\usepackage[table]{xcolor}
\usepackage{adjustbox}  
\usepackage{natbib}
\usepackage{orcidlink}

\usepackage[scaled]{helvet} 
 
\usepackage{mathpazo} 
\title{\textbf{Barlow-Swin}: Toward a novel siamese-based segmentation architecture using Swin-Transformers}

\author{
  Morteza Kiani Haftlang*\orcidlink{0009-0000-6147-200X} \\
  HSLU, Lucerne University of Applied Sciences and Arts, Lucerne, Switzerland \\
  \texttt{kianihaftlang@outlook.com} \\
  \And
  Mohammadhossein Malmir\orcidlink{0000-0003-0610-7899} \\
  Technical University of Munich, Munich, Germany \\
  \texttt{hossein.malmir@tum.de} \\
  \And
  Foroutan Parand\orcidlink{0000-0000-0000-0000} \\
  University College London (UCL), London, United Kingdom \\
  \texttt{f.parand@ucl.ac.uk} \\
  \And
  Umberto Michelucci\orcidlink{0000-0002-6060-5365} \\
  TOELT LLC AI lab, Winterthur, Switzerland \\
  \texttt{umberto.michelucci@toelt.ai} \\
  \And
  Safouane EL GHAZOUALI*\orcidlink{0000-0002-5403-3911} \\
  TOELT LLC AI lab, Winterthur, Switzerland \\
  \texttt{safouane.elghazouali@toelt.ai} \\
}

\begin{document}
\maketitle


\begin{abstract} 

    Medical image segmentation is a critical task in clinical workflows, particularly for the detection and delineation of pathological regions. While convolutional architectures like U-Net have become standard for such tasks, their limited receptive field restricts global context modeling. Recent efforts integrating transformers have addressed this, but often result in deep, computationally expensive models unsuitable for real-time use. In this work, we present a novel end-to-end lightweight architecture designed specifically for real-time binary medical image segmentation. Our model combines a Swin Transformer-like encoder with a U-Net-like decoder, connected via skip pathways to preserve spatial detail while capturing contextual information. Unlike existing designs such as Swin Transformer or U-Net, our architecture is significantly shallower and competitively efficient. To improve the encoder's ability to learn meaningful features without relying on large amounts of labeled data, we first train it using Barlow Twins, a self-supervised learning method that helps the model focus on important patterns by reducing unnecessary repetition in the learned features. After this pretraining, we fine-tune the entire model for our specific task. Experiments on benchmark binary segmentation tasks demonstrate that our model achieves competitive accuracy with substantially reduced parameter count and faster inference, positioning it as a practical alternative for deployment in real-time and resource-limited clinical environments. The code for our method is available at Github repository: \url{https://github.com/mkianih/Barlow-Swin}.
        
\end{abstract}

\section{Introduction}
Medical image segmentation is a fundamental task in computer-aided diagnosis, guiding clinicians in identifying and delineating critical anatomical or pathological structures \cite{ref:ronneberger2015u}. Conventional convolutional neural network (CNN) architectures, most notably the U-Net family \cite{ref:ronneberger2015u}, have achieved remarkable success in this domain. However, the limited receptive field of standard convolutions often restricts global context modeling, which can be critical for segmenting complex or subtle regions in medical images \cite{ref:gang2025convolutional,ref:li2021msu}.

In parallel, Transformers \cite{ref:transformer_original}\cite{ ref:vit_original} have recently shown promise in vision tasks, capturing both short- and long-range dependencies by virtue of the self-attention mechanism \cite{ref:schlemper2019attention}. Merging these two paradigms, hybrid U-Net-like models with Transformer encoders, such as TransUNet \cite{ref:chen2021transunet} and Swin-Unet \cite{ref:swinunet}, are an emerging trend for high-accuracy medical segmentation tasks.

Self-supervised learning (SSL) further addresses a key obstacle in medical imaging: the shortage of large-scale labeled data \cite{ref:jin2023label}. By leveraging redundancy and data augmentations, SSL methods can pretrain encoders to extract robust, generalizable features without extensive annotations. Among the various SSL approaches, Barlow Twins \cite{ref:zbontar2021barlow} has gained attention for its less complex and effective redundancy reduction objective. A prior work integrated Barlow Twins into a U-Net backbone, demonstrating improved encoder representations \cite{ref:chaitanya2020contrastive, ref:2021btunet}. Nevertheless, this line of research remains underexplored, particularly with Transformer-based encoders.

In this paper, we introduce \emph{Barlow-Swin}, a novel hybrid end-to-end lightweight architecture that benefits from a Swin Transformer-like encoder \cite{ref:liu2021swin} with a U-Net-like decoder. Encoder is first pretrained via Barlow Twins on the same medical imaging datasets. By preserving high-level global context from the shifted window-based self-attention mechanism \cite{ref:li2018hdenseunet} and fusing it with low-level spatial details through skip connections, our method aims to produce high-fidelity segmentation masks. Barlow Twins pretraining is used to enhance representation quality without requiring extensive labeled sets. In contrast to deeper Swin Transformer architectures like SwinUnet and TransUnet \cite{ref:swinunet} \cite{ref:chen2021transunet} , we opt for a shallower configuration to ensure computational efficiency, making it more attractive for real-time or resource-constrained environments.

We evaluate Barlow-Swin on four public medical image datasets (BCCD, BUSIS, ISIC2016, and a retinal dataset), each covering a diverse set of segmentation challenges. Our experiments compare the proposed architecture with classical U-Net baselines, the BT-UNet \cite{ref:2021btunet}, as well as other modern designs such as YOLOv8 \cite{ref:yolov8}, Segment Anything (SAM) \cite{ref:kirillov2023segment}, and HoverNet \cite{ref:graham2019hover}. Barlow-Swin achieves competitive or superior performance in terms of Dice coefficient and other metrics, underscoring the benefit of self-supervised Swin encoders for medical image segmentation. Qualitative visualizations are also provided, illustrating that Barlow-Swin can more accurately capture boundary details relative to purely convolutional architectures.

The remainder of this paper is organized as follows. Section 2 surveys the literature on U-Net variants, vision transformers in medical applications, and self-supervised learning. Section 3 outlines the Barlow-Swin architecture and SSL pretraining strategy. Section 4 describes the datasets, training protocols, and experimental results. Section 5 presents a discussion of our findings. Finally, Section 6 concludes the paper.


\section{Related Works}

Medical image segmentation has a long history, with classical methods preceding the advent of deep neural networks. Early approaches included thresholding, region growing, active contours (snakes) ~\cite{ref:kass1988snakes}, and atlas-based segmentation~\cite{ref:ashburner2000voxel}\cite{ref:pham2000current}, which relied heavily on handcrafted features and prior anatomical knowledge. While effective in certain scenarios, these methods often struggled with variability in anatomy, noise, and imaging artifacts.

The field underwent a paradigm shift with the introduction of deep learning, particularly fully convolutional networks (FCNs) and the U-Net family~\cite{ref:ronneberger2015u,ref:cicek20163dunet}. U-Net and its variants, such as U-Net++~\cite{ref:zhou2018unetpp}, ResUNet~\cite{ref:isensee2021nnunet}, and UNet 3+~\cite{ref:huang2020unet3plus}, utilize a symmetric encoder-decoder architecture with skip connections, achieving state-of-the-art performance on a wide range of segmentation tasks across modalities including MRI~\cite{ref:menze2015multimodal}, CT~\cite{ref:bilic2019liver}, and retinal fundus imaging~\cite{ref:staal2004ridge}. These architectures have become the de facto standard due to their ability to learn hierarchical features and facilitate end-to-end training.

Despite their success, convolutional neural networks are inherently limited by the local receptive fields of their convolutional kernels. This restricts their capacity to model long-range dependencies and global context, which are often crucial for accurately segmenting complex or subtle anatomical structures~\cite{ref:chen2021transunet,ref:zhou2023unetformer}. For example, distinguishing between adjacent organs or identifying diffuse pathological regions may require integrating information from distant parts of the image—something classical CNNs struggle to achieve. Several studies have highlighted that this limitation can lead to suboptimal performance, particularly in cases where global context is essential for correct delineation~\cite{ref:chen2021transunet}.

To address these challenges, recent research has explored the integration of self-attention mechanisms and transformer-based architectures, which are inherently better at modeling global dependencies. This evolution has led to the development of hybrid models such as TransUNet~\cite{ref:chen2021transunet} and other vision transformer-based approaches, which are discussed in detail in the following section.

\subsection{Transformers in Medical Imaging}

Transformers, originally developed for NLP tasks~\cite{ref:transformer_original}, have been increasingly adopted in vision applications due to their ability to model global dependencies via self-attention~\cite{ref:gu2019cenet}. The Vision Transformer (ViT)~\cite{ref:vit_original,ref:parmar2018imagetransformer} introduced the idea of splitting an image into fixed-size patches and treating them as tokens for sequential processing. However, its flat, non-hierarchical architecture limits scalability to the high-resolution images common in medical domains, as the computational and memory requirements grow rapidly with image size.

To address these limitations, the Swin Transformer~\cite{ref:liu2021swin} was proposed as a hierarchical alternative. It applies self-attention within local windows and shifts the windows between layers to achieve efficient cross-window communication. This design enables linear scaling with image size while retaining the benefits of self-attention. Swin Transformer has demonstrated strong performance on classification, detection, and segmentation benchmarks, and has recently been applied to medical images with encouraging results~\cite{ref:swinunet}. Nevertheless, Swin-based models can still be relatively deep and computationally demanding, and may not fully leverage the spatial inductive biases that are beneficial for fine-grained medical segmentation. These gaps have motivated the development of hybrid models that combine the strengths of CNNs and Transformers, which are discussed in the following section.

\subsection{CNN-Transformer Hybrids for Segmentation}

Several architectures have attempted to fuse the local feature extraction strengths of convolutional networks with the global context modeling capabilities of Transformers~\cite{ref:li2018hdenseunet,ref:he2016resnet}. A prominent example is TransUNet~\cite{ref:chen2021transunet}, which employs a CNN-based encoder to extract low-level features, followed by a Transformer (ViT) core operating at the bottleneck layer to model long-range dependencies, before upsampling with a U-Net-like decoder. In TransUNet, the ViT module processes the spatially reduced feature map as a sequence of tokens, enabling global context integration at the network’s deepest stage. This hybrid design achieves a balance between local detail retention and global context awareness. However, TransUNet inherits computational inefficiencies from its ViT bottleneck, particularly the quadratic complexity of self-attention, which can limit its scalability for very high-resolution medical images.

Swin-Unet~\cite{ref:swinunet}, another hybrid model, replaces the ViT bottleneck with a Swin Transformer backbone. It maintains the U-Net’s overall structure but uses Swin blocks in the encoder stages, enabling better multi-scale representation and long-range modeling. The hierarchical nature of Swin Transformer makes Swin-Unet more scalable for dense prediction tasks such as segmentation. Despite these advances, both TransUNet and Swin-Unet remain relatively deep and computationally intensive, which can be a barrier for real-time or resource-constrained clinical applications. This motivates the exploration of lighter-weight, more efficient hybrid designs, as introduced in this work.

\subsection{Self-Supervised Learning in Medical Imaging}

Self-supervised learning (SSL) has emerged as a promising technique to address the scarcity of annotated medical images. Methods such as SimCLR~\cite{ref:simclr}, MoCo~\cite{ref:moco}, and Barlow Twins~\cite{ref:zbontar2021barlow} enable the learning of invariant and semantically meaningful representations by comparing different augmented views of the same input~\cite{ref:caron2018deepcluster,ref:caron2020swav}. In this paradigm, an encoder is first trained to maximize agreement between these views, encouraging the model to focus on robust, generalizable features.

Barlow Twins, in particular, reduces redundancy between latent representations by encouraging the cross-correlation matrix of twin views to approach the identity matrix. This approach has proven effective in various domains and requires neither negative pairs nor momentum encoders. In the context of segmentation, it has been integrated with CNN-based architectures (e.g., BT-UNet~\cite{ref:2021btunet}) but remains underexplored in Transformer-based pipelines.

In this work, we present a novel and efficient architecture designed with the specific challenges of medical image analysis in mind. Inspired by the strengths of Swin Transformer, U-Net, and Barlow Twins, we developed a lightweight model that uses just three Swin stages and a simplified decoder to strike a balance between performance and computational efficiency. What sets our approach apart is the use of a self-supervised pretraining stage based on Barlow Twins, which helps the model learn meaningful features from unlabeled data before fine-tuning on the target task. By thoughtfully combining and adapting these components, our method offers a practical and effective solution rather than a straightforward integration of existing techniques~\cite{ref:grill2020byol}. 

\section{Methodology}

Our proposed Barlow-Swin approach for medical image segmentation integrates a Swin Transformer encoder (pretrained via Barlow Twins) with a lightweight U-Net-like decoder for segmentation outputs. We first present the overall architecture(Fig.~\ref{fig:barlow-swin-chart}), then expand on the mathematics behind the Barlow Twins objective, outline how our Swin Transformer and U-Net modules are constructed, and finally detail the loss functions used for final segmentation training.

\subsection{Overall Architecture}

\begin{figure*}[ht]
    \centering
    \includegraphics[width=\textwidth]{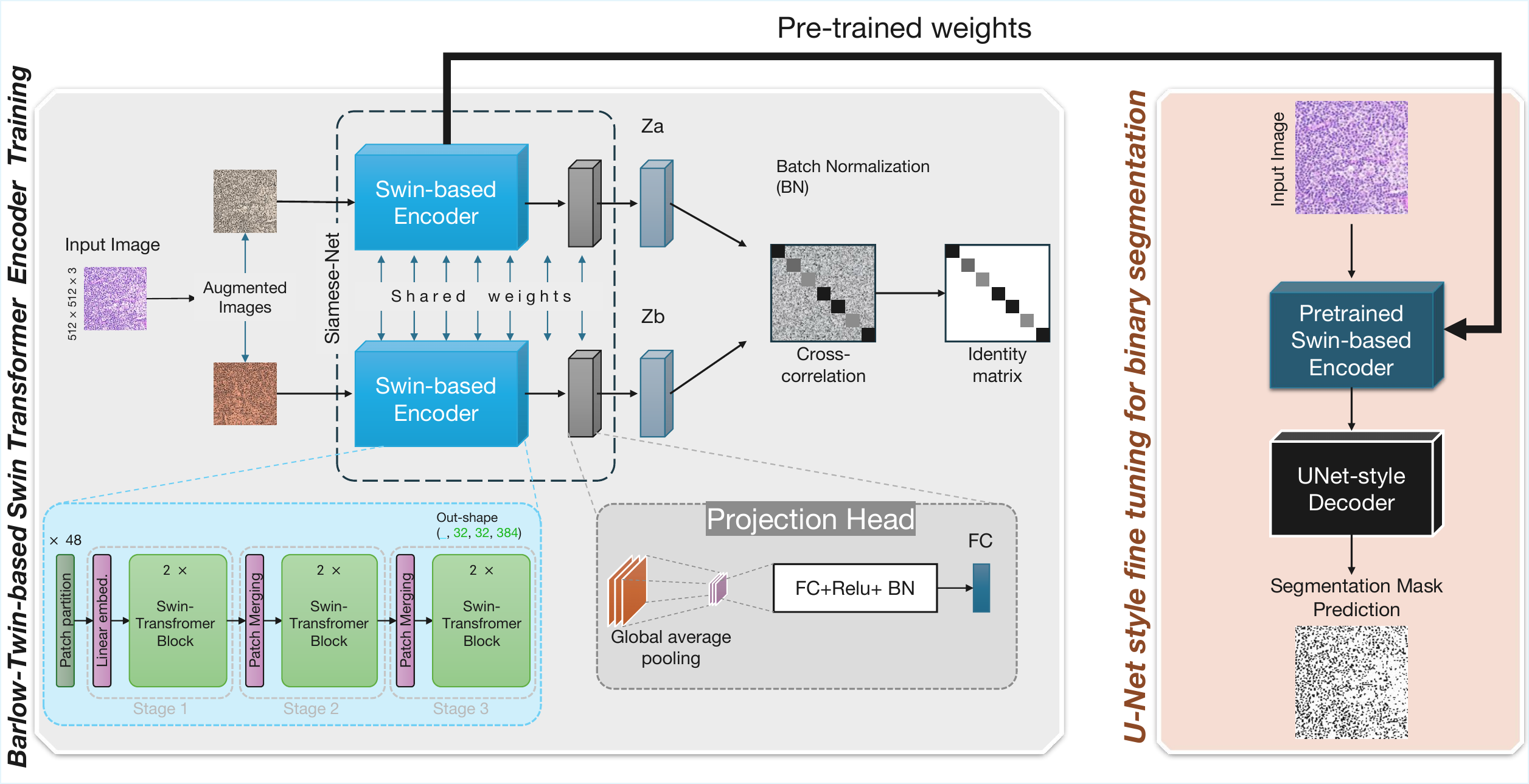}
        \caption{Overview of the proposed Barlow-Swin architecture. The model consists of two key stages: (1) a self-supervised pretraining phase, where the Swin Transformer-based encoder is trained using the Barlow Twins redundancy reduction objective on unlabeled medical images; and (2) a fine-tuning phase, where the pretrained encoder is integrated into a U-Net-style decoder for supervised binary segmentation. Skip connections between the encoder and decoder help preserve spatial detail while maintaining global context from the transformer backbone. 
    }
    \label{fig:barlow-swin-chart}
\end{figure*}

Our proposed framework, Barlow-Swin, consists of two key phases: (1) a self-supervised pretraining stage that leverages Barlow Twins to learn domain-relevant representations without requiring manual annotations, and (2) a fine-tuning stage that uses the pretrained encoder in a Swin-UNet architecture for binary medical image segmentation.

Given a dataset $\mathcal{X} = \{\mathbf{x}_i\}_{i=1}^{N}$ of unlabeled images and a set of corresponding binary masks $\mathcal{Y} = \{\mathbf{y}_i\}_{i=1}^{M}$ with $M \leq N$, the learning framework is decomposed as follows:

\textbf{Self-Supervised Representation Learning:}
We first train an encoder $f_\theta$ on the unlabeled dataset $\mathcal{X}u$ using the Barlow Twins objective. This objective encourages the encoder to produce representations $\mathbf{z} = f\theta(\mathbf{x})$ that are invariant to a range of appearance- and geometry-preserving transformations, such as random cropping, flipping, intensity jittering, and elastic deformation. These augmentations are designed to mimic the natural variability in medical images (e.g., anatomical differences, imaging artifacts), and the invariance constraint ensures that the learned representations focus on semantically meaningful and robust features rather than superficial variations. As a result, the encoder captures both global context (via the Swin Transformer backbone) and local structure, which are crucial for downstream segmentation tasks.

\textbf{Segmentation Fine-Tuning:}
The pretrained encoder $f_\theta$ is then integrated into a U-Net-like architecture with a lightweight decoder $g_\phi$, forming the segmentation model $g_\phi(f_\theta(\mathbf{x}))$. This model is trained on the labeled set $(\mathcal{X}_l, \mathcal{Y}l)$ using a supervised segmentation loss $\mathcal{L}{\mathrm{seg}}$. The skip connections between encoder and decoder stages allow for the fusion of low-level spatial details with high-level semantic context, further enhancing segmentation accuracy. 

Formally, the full training process minimizes the following two objectives in sequence:
\begin{itemize}
    \item Phase I (Barlow Twins Pretraining):
    \begin{equation}
        \min_{\theta} \; \mathcal{L}_{\mathrm{BT}}(f_\theta; \mathcal{X}),
    \end{equation}
    where $\mathcal{L}_{\mathrm{BT}}$ is the redundancy reduction loss defined in Eq.~\eqref{eq:bt_loss}.
    
    \item Phase II (Supervised Fine-Tuning):
    \begin{equation}
        \min_{\theta, \phi} \; \mathcal{L}_{\mathrm{seg}}(g_\phi(f_\theta(\mathbf{x})), \mathbf{y}),
    \end{equation}
    where $\mathcal{L}_{\mathrm{seg}}$ is the segmentation loss from Eq.~\eqref{eq:seg_loss}, optimized using the available ground truth masks.
\end{itemize}

This two-step strategy enables the encoder to learn strong contextual features from unlabeled medical data, while the decoder focuses on recovering detailed structures during segmentation. Compared to purely supervised training, our approach improves generalization, especially in settings with limited labeled examples. A schematic diagram of the full framework is shown in Fig.~\ref{fig:barlow-swin-chart}.

\subsection{Full Model Architecture}
\begin{figure*}[ht]
    \centering
    \includegraphics[width=\textwidth]{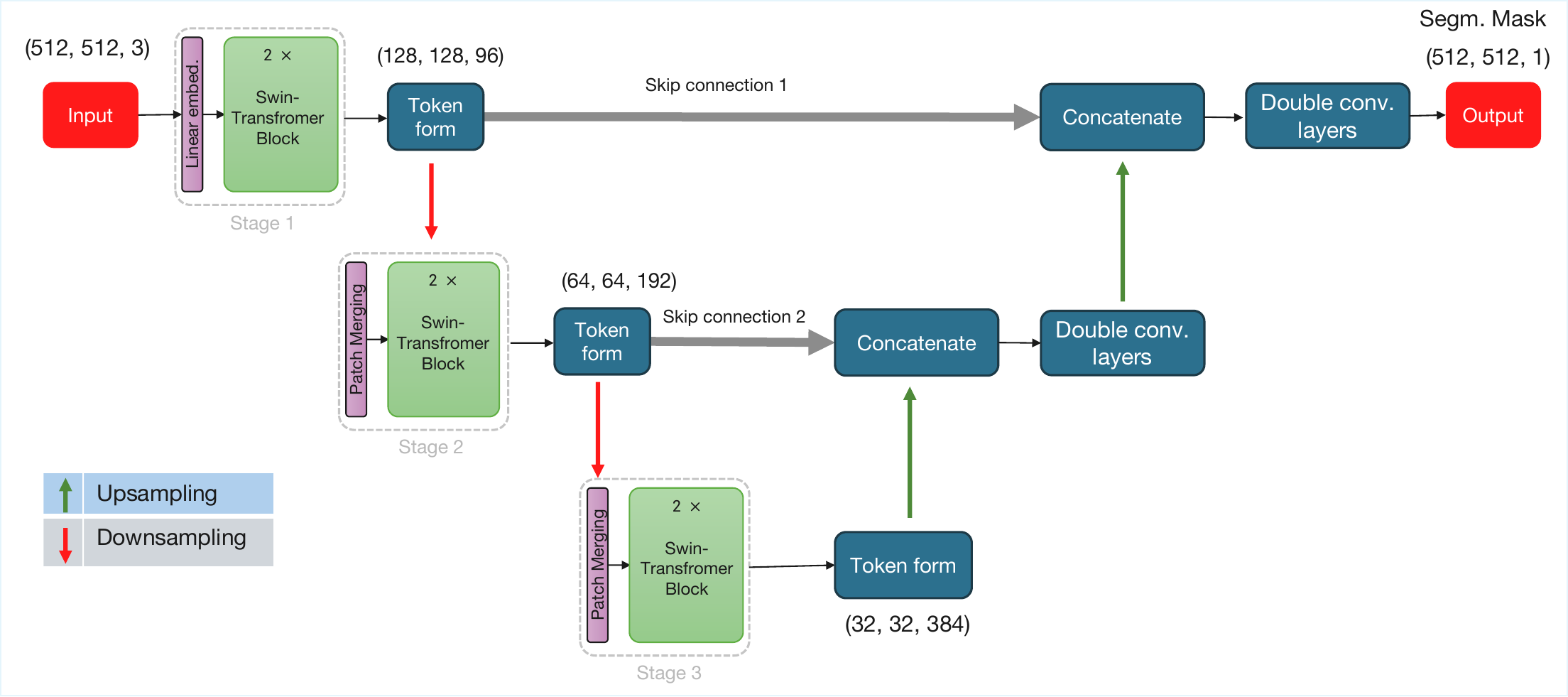}
    \caption{End-to-end architecture of the Barlow-Swin model. Input images are passed through a Swin Transformer-like encoder to extract multi-scale contextual features. A decoder with upsampling and skip connections reconstructs segmentation masks. The model is pretrained with Barlow Twins and then fine-tuned on labeled datasets.}
    \label{fig:full-model-architecture}
\end{figure*}

Our model follows a typical encoder-decoder paradigm, depicted in Fig.~\ref{fig:full-model-architecture}. The encoder is a Swin Transformer-like (Fig.~\ref{fig:encoder_diagram}) that accepts an image $\mathbf{I}$ of size $H \times W$ and partitions it into non-overlapping $4\times4$ patches. These patches are then embedded into latent tokens and passed through multiple Swin Transformer stages that progressively reduce spatial resolution and increase channel dimensionality. By the final stage, the encoder produces rich feature maps with both local and semi-global context. Meanwhile, a U-Net-style decoder recovers a full-resolution segmentation map from these deep feature representations, leveraging skip connections from intermediate Swin blocks to reintroduce fine-grained spatial details.

\subsection{Architectural Components}

\subsubsection{Swin Transformer Overview}
The Swin Transformer \cite{ref:liu2021swin} is a hierarchical vision transformer that introduces a shifted window mechanism to efficiently model both local and global dependencies. The architecture operates in stages, where each stage consists of multiple Swin Transformer blocks. At each block, self-attention is applied within fixed-size non-overlapping windows, allowing computational efficiency. To capture cross-window context, subsequent blocks shift the window partitioning, enabling communication between neighboring regions. The hierarchical nature of the model allows it to progressively downsample feature maps, akin to CNNs, while retaining the ability to model long-range spatial relations.

\begin{figure*}[ht]
    \centering
    \includegraphics[width=\textwidth]{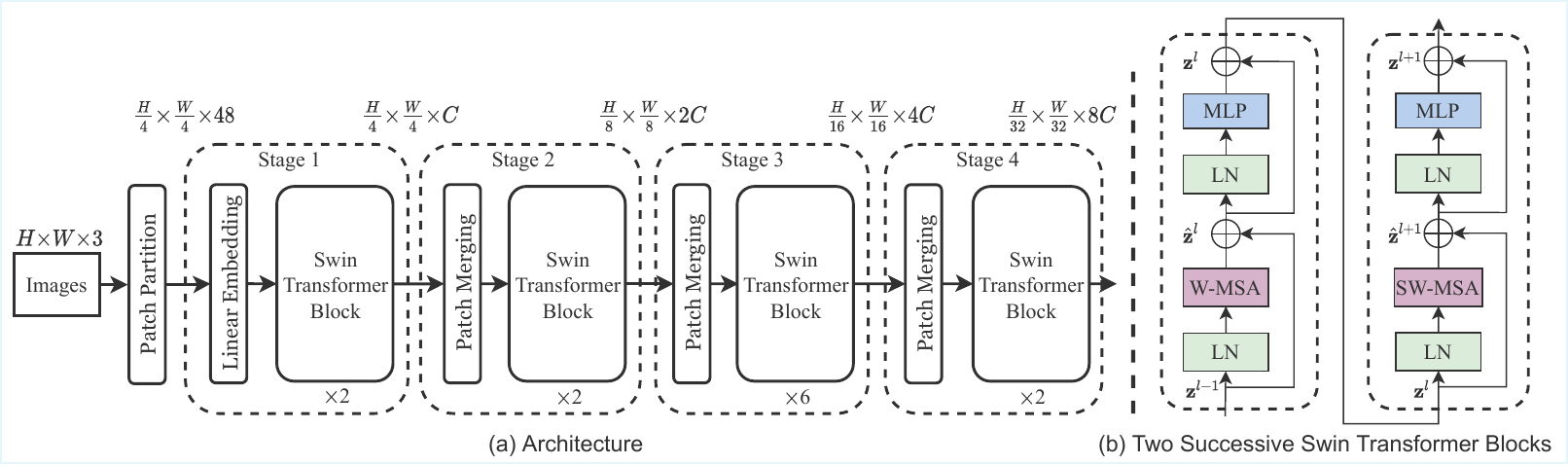}
    \caption{Structure of the Swin Transformer encoder used in original paper. (a) The architecture of a Swin Transformer (Swin-T); (b) two successive Swin Transformer Blocks. The encoder partitions the input image into $4 \times 4$ patches and processes them through hierarchical Swin blocks with shifting window attention. This approach captures both local and global features efficiently. 
}
    \label{fig:encoder_diagram}
\end{figure*}

\subsubsection{Swin-Like Encoder}
\begin{figure*}[ht]
    \centering
    \includegraphics[width=\textwidth]{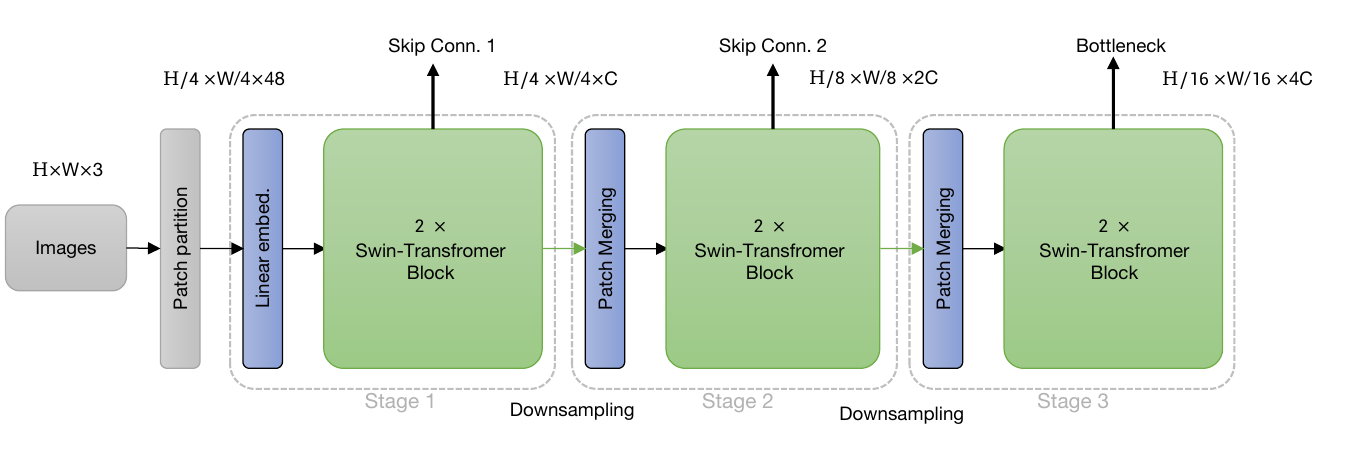}
    \caption{Detailed visualization of the three-stage Swin Transformer encoder architecture. Each stage downsamples spatial resolution and increases feature dimensionality. Skip connections are extracted from stages 1 and 2 for decoder fusion. }
    \label{fig:encoder_diagram_proposed}
\end{figure*}

Our encoder adopts a three-stage Swin Transformer design(Fig.~\ref{fig:encoder_diagram_proposed}) comparing to that of original paper(Fig.~\ref{fig:encoder_diagram}) with four stages. Starting from an input of size $512\times512$, it first partitions the image into non-overlapping $4\times4$ patches, resulting in $128\times128$ patch tokens. These are passed through:
\begin{itemize}
    \item Stage 1: two Swin Transformer blocks with $96$ channels,
    \item Stage 2: Patch Merging followed by two Swin blocks with $192$ channels,
    \item Stage 3: another Patch Merging and two Swin blocks with $384$ channels.
\end{itemize}
Each stage halves the spatial resolution and increases the number of channels. This produces hierarchical feature maps at scales: $(128,128,96)$, $(64,64,192)$, and $(32,32,384)$, which are then used as skip connections.

\subsubsection{U-Net Overview}

U-Net \cite{ref:ronneberger2015u} is a widely adopted convolutional neural network architecture specifically designed for biomedical image segmentation. Its hallmark U-shaped structure comprises two main paths: a contracting encoder that captures contextual information through progressive downsampling, and an expansive decoder that restores spatial resolution for precise localization.

The encoder path operates similarly to classical convolutional neural networks (CNNs), applying sequential convolutional layers followed by max-pooling operations. This process reduces spatial dimensions while increasing feature depth, allowing the model to learn hierarchical representations from low-level edges to complex structures. 

What makes U-Net fundamentally different from earlier segmentation models, such as Fully Convolutional Networks (FCN) \cite{ref:long2015fcn} and SegNet \cite{ref:badrinarayanan2017segnet}, is the introduction of skip connections between the encoder and decoder at each resolution level. These connections concatenate corresponding feature maps, ensuring that fine-grained spatial information lost during downsampling is directly restored during upsampling. This design is particularly beneficial for medical imaging tasks, where boundary precision and the retention of small structures are critical.

The standard U-Net consists of four downsampling stages, each halving the spatial resolution and doubling the number of feature channels. The decoder path mirrors this hierarchy, employing up-convolutions (transposed convolutions) or interpolation-based upsampling to incrementally restore the image resolution while refining features. Each upsampling step is followed by concatenation with the corresponding encoder output and convolutional layers to reconstruct segmentation maps with high fidelity.

Despite its remarkable success across diverse medical imaging tasks, including cell segmentation, tumor boundary detection, and organ delineation, U-Net and its convolutional derivatives are inherently limited by the local receptive field of convolutional operations. This limitation hinders the model’s capacity to capture long-range dependencies or global context, which is crucial for accurately segmenting large anatomical regions or distinguishing spatially distant yet contextually related structures.

These limitations have led to a growing interest in integrating transformer-based architectures into segmentation pipelines. Transformers, with their self-attention mechanisms, naturally capture both local and global dependencies, addressing the context-awareness gap left by CNN-based models.

\subsubsection{U-Net-like Decoder}

\begin{figure*}[ht]
    \centering
    \includegraphics[width=\textwidth]{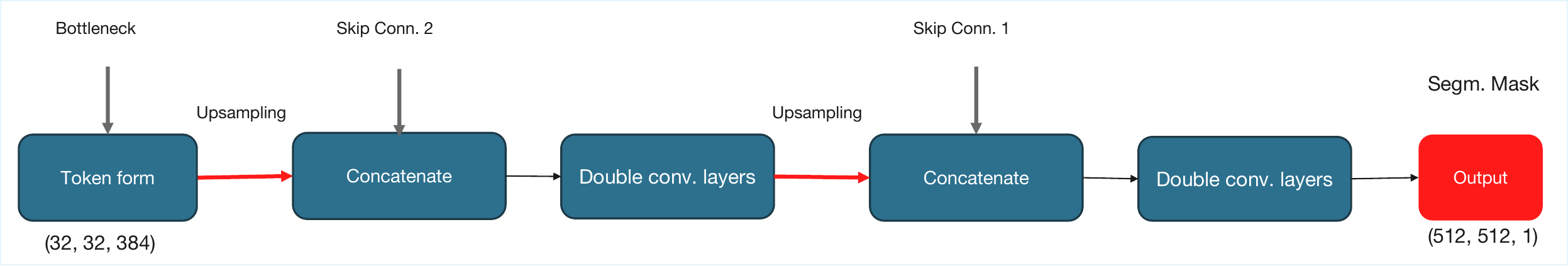}
    \caption{Schematic of the U-Net-like decoder used in Barlow-Swin. The decoder consists of a sequence of upsampling layers, depthwise-separable convolutions, and skip connections from the Swin Transformer encoder. A final $1 \times 1$ convolution produces the binary segmentation mask.}
    \label{fig:decoder_diagram}
\end{figure*}

The decoder in our proposed Barlow-Swin architecture in Fig.~\ref{fig:decoder_diagram} is inspired by the classical U-Net’s expansive path but is adapted to complement the Swin Transformer-based encoder. While the encoder leverages hierarchical shifted-window self-attention mechanisms to capture global context and multi-scale features, the decoder focuses on progressively recovering spatial resolution and refining local details to produce high-fidelity segmentation masks.

Unlike the encoder, which is purely transformer-based, the decoder retains convolutional inductive biases characteristic of U-Net \cite{ref:ronneberger2015u}. This design choice allows the model to efficiently combine the global semantic information from the Swin Transformer with the precise spatial information carried through skip connections.

Specifically, the decoder consists of a sequence of upsampling stages, each of which performs the following steps:
\begin{itemize}
    \item The feature map is upsampled using \texttt{UpSampling2D}, doubling its spatial dimensions.
    \item The upsampled feature is concatenated with the corresponding skip connection from the Swin Transformer encoder. These skip connections inject high-resolution features from earlier encoder layers, mitigating the spatial information loss inherent to downsampling.
    \item The concatenated features are processed through a double convolution block composed of two depthwise-separable convolution layers, each followed by batch normalization and ReLU activation. A residual connection with a $1 \times 1$ convolution stabilizes training and enhances representational capacity.
\end{itemize}

This decoder reconstructs the resolution in the following sequence:
\[
\begin{aligned}
(32, 32, 384)\; &\to\;(64, 64, 192)\;\to\;(128, 128, 96) \\
               &\to\;(256, 256, 64)\;\to\;(512, 512, 32)
\end{aligned}
\]

At the final stage, a $1 \times 1$ pointwise convolution reduces the channel dimension from 32 to 1, yielding the final segmentation mask with spatial dimensions $(512, 512, 1)$. This is followed by a sigmoid activation, producing pixel-wise probabilities suitable for binary segmentation tasks.

By coupling the self-attention-driven Swin Transformer encoder with a lightweight U-Net-like convolutional decoder, the architecture achieves an effective balance between capturing global context and preserving fine-grained spatial precision. This hybrid design proves particularly advantageous for medical image segmentation, where both large-scale context (e.g., anatomical structure) and local boundary details are critical.

\subsubsection{Skip Connections}
To retain fine-grained details from early layers, our model includes two skip connections:
\begin{itemize}
    \item From Stage 1 encoder output $(128,128,96)$ to the second decoder block,
    \item From Stage 2 encoder output $(64,64,192)$ to the first decoder block.
\end{itemize}
These features are concatenated with the decoder features before convolution and activation. The deepest encoder output $(32,32,384)$ is used as the decoder’s initial input, but not as a skip connection.

\subsubsection{Output Activation}
The final prediction layer is a $1\times1$ convolution that maps decoder features to a single-channel mask. We use the \texttt{sigmoid} activation function to produce probabilities in the range $[0,1]$ for each pixel. Notably, we do \emph{not} apply a softmax since our task is binary segmentation, and sigmoid suffices for modeling foreground vs. background.

\subsection{Barlow Twins Pretraining}
\begin{figure*}[ht]
    \centering
    \includegraphics[width=\textwidth]{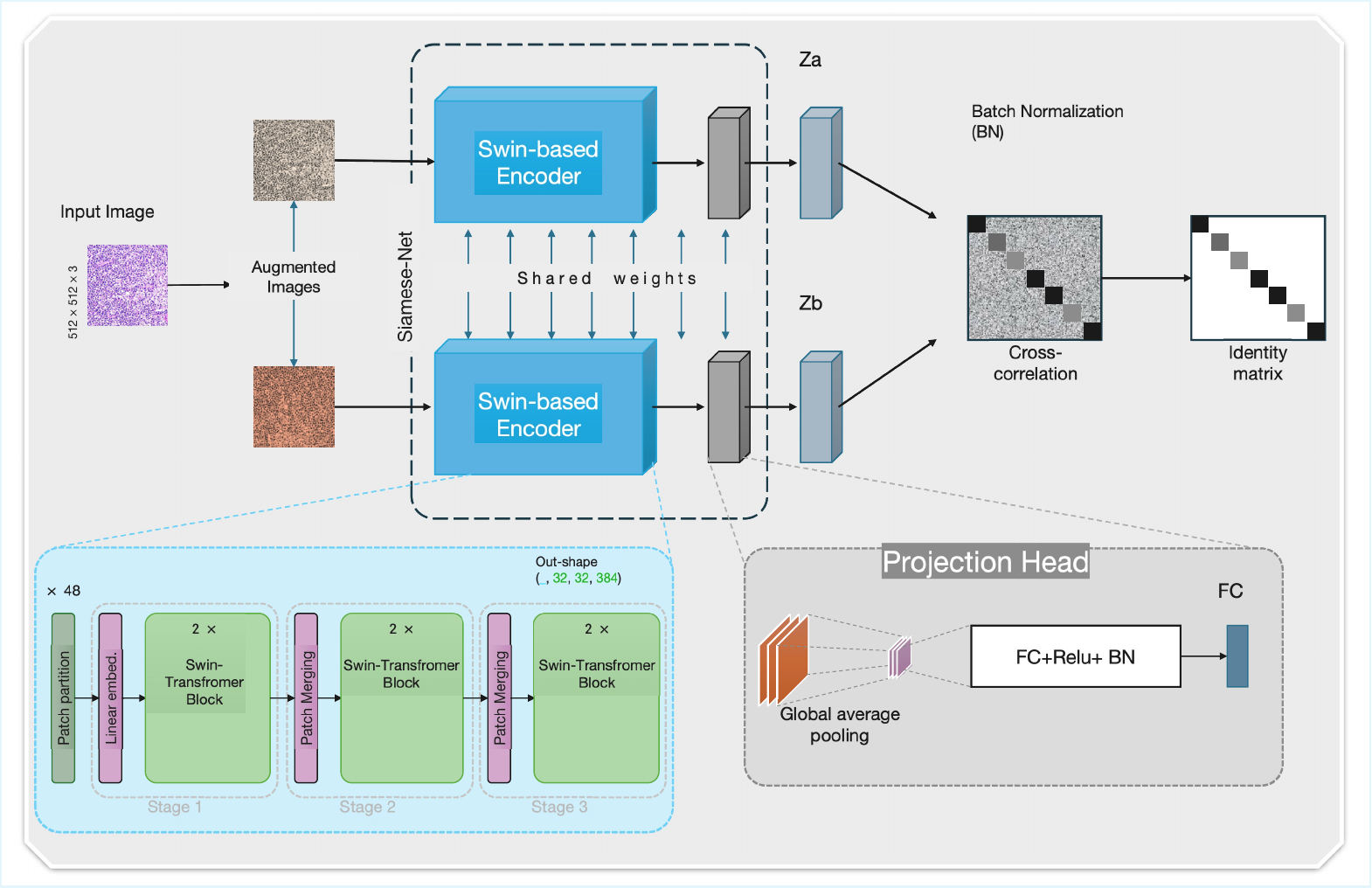}
    \caption{Illustration of the Barlow Twins self-supervised pretraining framework applied to the Swin Transformer encoder in our architecture. The top portion of the figure depicts the general Barlow Twins pipeline, where two augmented views of the same input are passed through a shared encoder to produce latent representations. These are then compared using a cross-correlation-based redundancy reduction loss. The bottom portion illustrates our adaptation, in which a Swin Transformer-based encoder, designed for hierarchical vision representation, replaces conventional CNN-based encoders commonly used in classical Barlow Twins implementations \cite{ref:zbontar2021barlow}. This design allows our encoder to capture both local and global contextual information while benefiting from self-supervised learning. The pretraining objective encourages the encoder to produce invariant yet non-redundant features across different augmentations, enhancing its generalization for downstream medical image segmentation tasks without requiring manual labels.}
    \label{fig:barlowtwin_framework}
\end{figure*}

A key aspect of our design is the self-supervised pretraining of the Swin encoder using Barlow Twins as shown in  Fig.~\ref{fig:barlowtwin_framework}  \cite{ref:zbontar2021barlow}. This step aims to address the limited availability of annotated medical data by learning useful feature representations from unlabeled samples.

\paragraph{Dual Views.}
We take an unlabeled image $\mathbf{x}$ and apply two distinct augmentations, producing $\mathbf{x}^{(1)}$ and $\mathbf{x}^{(2)}$. These can include geometric transformations (random cropping, flipping, rotation), color jitter, and other domain-specific augmentations. 

\paragraph{Encoder Extraction.}
Each augmented view is processed by the Swin encoder $f_{\theta}$ to yield latent representations:
\begin{equation}
    \mathbf{z}^{(1)} \;=\; f_{\theta}(\mathbf{x}^{(1)}), 
    \quad 
    \mathbf{z}^{(2)} \;=\; f_{\theta}(\mathbf{x}^{(2)}).
\end{equation}
Here, $f_{\theta}$ denotes the shared encoder whose weights are updated to satisfy the Barlow Twins redundancy reduction objective.

\paragraph{Redundancy Reduction Loss.}
Following \cite{ref:zbontar2021barlow}, we compute a cross-correlation matrix $\mathbf{C}$ between $\mathbf{z}^{(1)}$ and $\mathbf{z}^{(2)}$. Let $\mathbf{z}^{(1)}_i$ and $\mathbf{z}^{(2)}_i$ represent the $i$-th dimension of $\mathbf{z}^{(1)}$ and $\mathbf{z}^{(2)}$ (after batch normalization). For a mini-batch of $B$ samples, 
\begin{equation}
    C_{ij} 
    \;=\;
    \frac{\sum_{b}^{B} \bigl(\mathbf{z}^{(1)}_{b,i}\bigr)\,\bigl(\mathbf{z}^{(2)}_{b,j}\bigr)}{
    \sqrt{\sum_{b}^{B} \bigl(\mathbf{z}^{(1)}_{b,i}\bigr)^{2}}
    \;\sqrt{\sum_{b}^{B} \bigl(\mathbf{z}^{(2)}_{b,j}\bigr)^{2}}
    },
\end{equation}

The objective loss function aims to make the diagonal elements $C_{ii}$ close to 1 (invariance across views) and the off-diagonal elements $C_{ij}$ where $i \neq j$ close to 0 (decorrelation). The Barlow Twins loss is defined as:

where $1 \leq i,j \leq d$ and $d$ is the dimensionality of the latent vector. We then minimize:
\begin{equation}
    \label{eq:bt_loss}
        \mathcal{L}_{\mathrm{BT}} = \sum_{i} (1 - C_{ii})^2 + \lambda \sum_{i} \sum_{j \neq i} C_{ij}^2 
\end{equation}
where the first term (\emph{on-diagonal}) encourages invariance (i.e., each neuron yields the same representation for both augmented views), and the second (\emph{off-diagonal}) reduces redundancy among different neurons. The hyperparameter $\lambda$ balances these two objectives.

After pretraining, we obtain weights $\theta$ that capture meaningful feature representations from unlabeled data. We subsequently initialize our Swin encoder with these weights and train the complete segmentation pipeline on limited annotated data.

\subsection{Barlow-Swin Components}

\noindent \textbf{Patch Partition:} This component divides the input image into non-overlapping patches to facilitate local processing in the transformer. Let $\mathbf{I} \in \mathbb{R}^{H \times W \times C}$ represent the input image, where $H = W = 512$ and $C = 3$. Given a window size $w = 4$, the image is partitioned into patches as:
\begin{equation}
  \text{patches} \;=\; \mathrm{extract\_patches}\!\bigl(\mathbf{I}, \text{size}=[w,w]\bigr),
\end{equation}
where each patch is flattened into a vector of length $w \times w \times C = 48$. These patches form the token embeddings for the subsequent transformer stages.

\noindent \textbf{Linear Embedding:} This stage projects each flattened patch into a higher-dimensional embedding space and injects positional information. Each patch vector $\mathbf{p}_i \in \mathbb{R}^{48}$ is projected using a learnable linear transformation:
\begin{equation}
  \mathbf{z}_i \;=\; \mathbf{W}\,\mathbf{p}_i + \mathbf{b}, 
  \quad
  \mathbf{e}_i = \mathbf{z}_i + \mathbf{E}(i),
\end{equation}
where $\mathbf{W}\in \mathbb{R}^{96\times48}$ and $\mathbf{b}\in\mathbb{R}^{96}$ are trainable parameters, and $\mathbf{E}(i)$ represents a positional embedding that maintains spatial information within the image grid.

\noindent \textbf{Patch Merging:} This step reduces spatial resolution while expanding the feature dimensionality. Given a feature map $\mathbf{x} \in \mathbb{R}^{H' \times W' \times C}$, it is divided into four sub-sampled tensors:
\begin{equation}
\begin{split}
  \mathbf{x}_0 &= \mathbf{x}[0{:}{:}2,\,0{:}{:}2], \\
  \mathbf{x}_1 &= \mathbf{x}[1{:}{:}2,\,0{:}{:}2], \\
  \mathbf{x}_2 &= \mathbf{x}[0{:}{:}2,\,1{:}{:}2], \\
  \mathbf{x}_3 &= \mathbf{x}[1{:}{:}2,\,1{:}{:}2].
\end{split}
\end{equation}
Each sub-block has shape $\left( \tfrac{H'}{2}, \tfrac{W'}{2}, C \right)$. These are concatenated along the channel axis:
\begin{equation}
  \mathbf{x}_{\text{merged}} 
  \in \mathbb{R}^{\tfrac{H'}{2} \times \tfrac{W'}{2} \times (4C)},
\end{equation}
and then a linear layer projects the channel dimension from $4C$ to $2C$, achieving spatial downsampling along with channel expansion for the next stage.

\noindent \textbf{Multi-Layer Perceptron (MLP):} The MLP introduces non-linearity and improves feature mixing within transformer blocks. It consists of two fully connected layers with GELU activation and dropout:
\begin{equation}
  \mathrm{MLP}(\mathbf{x}) 
  \;=\; 
  \mathrm{Dropout}\Bigl(
    \mathrm{Dense}_2\bigl(
      \mathrm{Dropout}\bigl(\mathrm{GELU}(\mathrm{Dense}_1(\mathbf{x}))\bigr)
    \bigr)
  \Bigr),
\end{equation}
where $\mathrm{Dense}_1$ expands the feature dimension (typically by a factor of 4) and $\mathrm{Dense}_2$ projects it back to the original size. The GELU (Gaussian Error Linear Unit) activation is defined as:
\begin{equation}
\mathrm{GELU}(x) = x \cdot \Phi(x),
\end{equation}
where $\Phi(x)$ is the standard Gaussian cumulative distribution function. GELU provides a smooth and effective alternative to ReLU, enabling better learning dynamics in transformers.

\noindent \textbf{Window Attention:} This component captures local dependencies within non-overlapping windows using self-attention. Given input tokens $\mathbf{x} \in \mathbb{R}^{B \times N \times C}$, where $B$ is the batch size, $N = \text{window\_size}^2$ is the number of tokens per window, and $C$ is the channel dimension, the attention mechanism operates as:
\begin{equation}
  [Q,\,K,\,V] = \mathrm{Dense}(\mathbf{x}),
\end{equation}
and the scaled dot-product attention is computed as:
\begin{equation}
  \mathrm{Attention}(Q,K,V) 
  = 
  \mathrm{softmax}\!\left( \frac{Q K^\top}{\sqrt{d_k}} + \mathrm{relative\_bias} \right) V.
\end{equation}
Here, $\mathrm{relative\_bias}$ accounts for the positional relationships within the window to enhance spatial sensitivity.

\noindent \textbf{DropPath (Stochastic Depth):} This is a regularization technique that randomly drops entire residual branches during training to improve model generalization. It is computed as:
\begin{equation}
    \mathbf{y} = \frac{\mathbf{x}}{1-p} \cdot \mathrm{Bernoulli}(1-p),
\end{equation}
where $p$ is the drop probability. By randomly skipping paths, it prevents overfitting and improves robustness.

\noindent \textbf{Swin Transformer Block:} This block models local context with window-based self-attention combined with feed-forward MLP layers, enhanced with residual connections and optional window shifting. Each block performs:
\begin{itemize}
    \item[1.] \textit{LayerNorm:} Normalize the input tensor.
    \item[2.] \textit{Window Partition and Shift:} Apply cyclic shifts if $\mathrm{shift\_size}>0$ to allow cross-window interaction.
    \item[3.] \textit{Window Attention:} Compute local self-attention within each windowed region.
    \item[4.] \textit{Reverse Shift:} Revert the shift to maintain correct spatial positioning.
    \item[5.] \textit{Residual Connections:} Apply residual connections over both the attention module and the MLP module, with DropPath:
    \[
      \mathbf{x} \leftarrow \mathbf{x} + \mathrm{DropPath}(\mathbf{x}'_{\mathrm{attn}})
    \]
    \[
      \mathbf{x} \leftarrow \mathbf{x} + \mathrm{DropPath}(\mathrm{MLP}(\mathrm{LayerNorm}(\mathbf{x})))
    \]
\end{itemize}
By stacking multiple Swin Transformer blocks with alternating shift sizes, the encoder captures both local and global context efficiently.

\noindent \textbf{UNet-Style Decoder:} The decoder is designed to progressively recover spatial resolution and generate the final segmentation map. It consists of:
\begin{itemize}
    \item \textbf{Double Convolution Block:} Two depthwise-separable $3\times3$ convolutions followed by Batch Normalization and ReLU activation, plus a $1\times1$ convolutional shortcut for residual learning.
    \item \textbf{Upsampling:} Each decoder stage upsamples the feature maps by a factor of 2 and concatenates them with the corresponding encoder skip connections.
    \item \textbf{Final Output Layer:} A $1\times1$ convolution reduces the feature map from $(512,512,32)$ to $(512,512,1)$, followed by a sigmoid activation function to produce the binary segmentation mask.
\end{itemize}
The decoding follows the resolution sequence:
(32,32,384) $\rightarrow$ (64,64,192) $\rightarrow$ (128,128,96) $\rightarrow$ (256,256,64) $\rightarrow$ (512,512,32) $\rightarrow$ (512,512,1)

\noindent \textbf{Loss Functions for Segmentation:} To guide the training of the segmentation model, we employ a hybrid loss function that combines Binary Cross-Entropy (BCE) with Dice Loss. This combination balances fine-grained pixel accuracy with robust region-level segmentation performance, particularly critical for medical image datasets that often exhibit class imbalance.

\noindent \textbf{Binary Cross-Entropy (BCE):} BCE evaluates pixel-wise prediction accuracy:
\begin{equation}
    \mathcal{L}_{\mathrm{BCE}}
    = -\frac{1}{N}\sum_{i=1}^{N}
    \left[y_i \log\hat{y}_i + (1 - y_i) \log(1 - \hat{y}_i)\right],
\end{equation}
where $N$ is the total number of pixels, $y_i$ is the ground truth label, and $\hat{y}_i$ is the predicted probability for pixel $i$.

\noindent \textbf{Dice Loss:} Dice Loss directly optimizes for the overlap between the predicted and ground truth masks:
\begin{equation}
    \mathrm{Dice}(\mathbf{y}, \hat{\mathbf{y}}) 
    = 
    \frac{2 \sum_i y_i \hat{y}_i}{
      \sum_i y_i + \sum_i \hat{y}_i + \epsilon
    },
\end{equation}
and the corresponding loss function is:
\begin{equation}
    \mathcal{L}_{\mathrm{Dice}} 
    = 1 - \mathrm{Dice}(\mathbf{y}, \hat{\mathbf{y}}).
\end{equation}

\noindent \textbf{Combined Loss:} The final segmentation loss combines BCE and Dice losses as:
\begin{equation}
    \mathcal{L}_{\mathrm{seg}} 
    = \alpha \mathcal{L}_{\mathrm{BCE}} + (1 - \alpha) \mathcal{L}_{\mathrm{Dice}},
    \label{eq:seg_loss}
\end{equation}
where $\alpha$ controls the balance between the two terms. This ensures the model not only achieves accurate pixel-level predictions but also maintains coherent and connected regions in the output mask.


\subsection{Datasets and Implementation Details}

We evaluate our proposed Barlow-Swin framework on four publicly available medical image segmentation datasets, each representing distinct modalities and segmentation challenges. The details of the datasets are as follows:

\begin{itemize}
    \item \textbf{BCCD} \cite{ref:bccd_dataset}: A microscopic blood cell dataset consisting of 364 annotated images. This dataset is primarily designed for segmentation and detection of red blood cells (RBCs), white blood cells (WBCs), and platelets in peripheral blood smears.
    
    \item \textbf{BUSIS} \cite{ref:busis_dataset}: The Breast Ultrasound Image Segmentation dataset contains 562 ultrasound images, each annotated with corresponding tumor masks. This dataset poses challenges due to noise, low contrast, and irregular tumor boundaries inherent to ultrasound imaging.
    
    \item \textbf{ISIC2016} \cite{ref:isic2016_dataset}: Provided by the International Skin Imaging Collaboration (ISIC), this dataset contains 900 dermoscopic images of skin lesions with corresponding binary segmentation masks. It focuses on delineating lesion boundaries for melanoma detection.
    
    \item \textbf{DRIVE Retina} \cite{ref:drive_dataset}: The Digital Retinal Images for Vessel Extraction (DRIVE) dataset consists of 40 high-resolution retinal fundus images annotated for vessel segmentation. This dataset is particularly challenging due to the thin and elongated nature of retinal blood vessels.
\end{itemize}

All images are resized to a uniform resolution of $512 \times 512$ to maintain architectural compatibility across models. Each dataset is partitioned into training (70\%), validation (15\%), and test (15\%) subsets using a fixed random seed to ensure reproducibility.

Data augmentations—including random rotations, horizontal and vertical flips, and color jitter—are applied during supervised training to improve generalization. The same augmentations are used to generate paired views for the Barlow Twins pretraining step, thereby promoting representation learning invariant to appearance transformations.

In the self-supervised pretraining phase, the Swin Transformer encoder is trained solely on the unlabeled images using the Barlow Twins redundancy reduction objective. After convergence, the pretrained encoder weights are transferred to the full segmentation model, which integrates the Swin encoder with a U-Net-like decoder. The entire network is subsequently fine-tuned end-to-end using the labeled data with the combined BCE-Dice loss function (Eq.~\eqref{eq:seg_loss}).

Training employs the Adam optimizer with an initial learning rate of $1 \times 10^{-4}$, a batch size of 8, and early stopping based on validation loss stagnation. Each model is trained for a maximum of 200 epochs unless early stopping criteria are met. In total, we conducted 24 experiments, covering six different model variants across the four datasets. Unless otherwise specified, all transformer hyperparameters, such as embedding dimension, window size, and number of heads, follow the original Swin Transformer design.

\subsection{Baselines}

To comprehensively evaluate the performance of Barlow-Swin, we compare it against several strong baselines, encompassing both classical CNN-based methods and recent transformer-based or detection-driven models:

\begin{itemize}
    \item \textbf{U-Net} \cite{ref:ronneberger2015u}: A widely used convolutional encoder-decoder architecture that leverages skip connections for precise biomedical image segmentation.
    
    \item \textbf{BT-UNet} \cite{ref:2021btunet}: A self-supervised variant of U-Net pretrained using the Barlow Twins objective. This model mirrors our pretraining pipeline but utilizes a convolutional encoder instead of a transformer-based one.
    
    \item \textbf{HoverNet} \cite{ref:graham2019hover}: Originally designed for simultaneous nuclear instance segmentation and classification. Despite its specialization in nuclear segmentation, it serves as a competitive baseline for dense prediction tasks.
    
    \item \textbf{YOLOv8Seg} \cite{ref:yolov8}: A real-time segmentation variant based on the YOLOv8 architecture. It balances speed and accuracy but is primarily designed for general-purpose object segmentation rather than dense medical segmentation.
    
    \item \textbf{YOLO-SAM} \cite{ref:kirillov2023segment}: A hybrid pipeline that combines YOLOv8 bounding box detection with the Segment Anything Model (SAM) for instance-agnostic segmentation. While highly versatile in zero-shot settings, it has not been specifically optimized for pixel-precise biomedical segmentation.
\end{itemize}

All baseline models are adapted to accept $512 \times 512$ inputs for a fair comparison. Unless a model is inherently pretrained (such as YOLO-SAM initialized with SAM's foundation weights), all models, including YOLOv8Seg and HoverNet, are fine-tuned end-to-end—not merely their segmentation heads—using the same training, validation splits, and augmentation pipeline as Barlow-Swin. This ensures that all models are trained under identical conditions and loss functions (combined BCE-Dice loss) to maintain fairness in evaluation.

Fine-tuning for the transformer-based Barlow-Swin and BT-UNet follows the same two-stage protocol: (1) self-supervised pretraining using unlabeled images with Barlow Twins, followed by (2) supervised fine-tuning on labeled segmentation masks. For YOLOv8Seg and YOLO-SAM, pretrained detection weights are used, but the segmentation modules are specifically fine-tuned on the medical datasets to align with the evaluation protocol of Barlow-Swin.

\subsection{Qualitative Comparison}

\begin{figure*}[htbp!]
  \centering
  \includegraphics[width=0.9\linewidth]{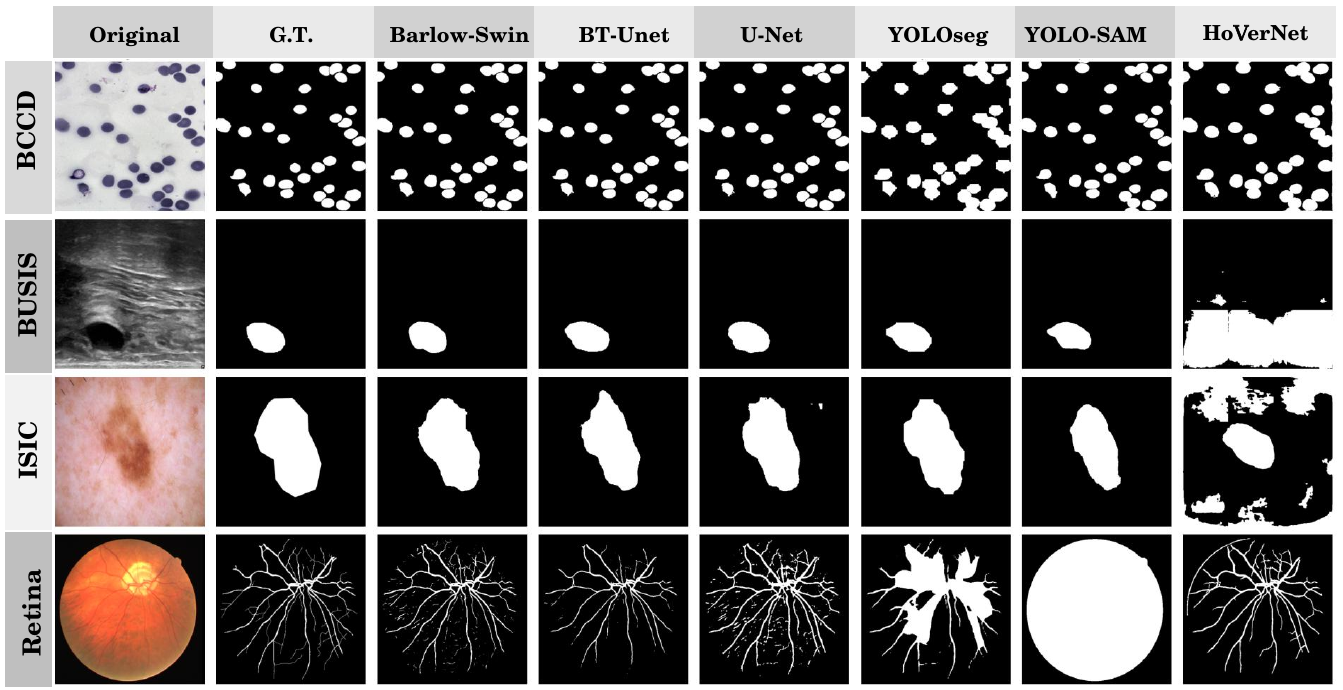} 
  \caption{Visual comparison of segmentation predictions across the four datasets. Each row displays an input image, ground truth mask, and predictions from several models, including U-Net, BT-UNet, YOLOv8-Seg, and the proposed Barlow-Swin. Barlow-Swin demonstrates superior boundary adherence and reduced false positives, particularly on vessel (Retina) and lesion (ISIC) images.}
  \label{fig:qualitative_predictions}
\end{figure*}

Fig.~\ref{fig:qualitative_predictions} presents a visual comparison of segmentation outputs across all models for one representative image from each dataset. Overall, Barlow-Swin consistently produces cleaner and more accurate segmentation results compared to the baseline models. In the Retina dataset, for example, it captures the thin and intricate vessel structures with higher continuity, where other models tend to miss finer branches or produce broken segments. On the ISIC skin lesion dataset, Barlow-Swin demonstrates competitively better boundary precision, particularly around irregular contours, and generates fewer false positives in the surrounding skin compared to convolution-based models like U-Net or BT-UNet.

Similarly, in the BUSIS ultrasound dataset, which is characterized by low contrast and noisy backgrounds, Barlow-Swin shows improved robustness by suppressing spurious artifacts and producing smoother, more reliable tumor boundaries. On the BCCD dataset, focused on blood cell segmentation, the model is able to delineate individual cells with clearer borders, reducing common errors such as boundary bleeding or over-segmentation that are often observed in models like HoverNet or YOLOv8-Seg.

These observations highlight that the combination of Swin Transformer's hierarchical attention with self-supervised pretraining via Barlow Twins enables the model to better balance global context with local detail. This translates into improved boundary adherence and artifact reduction across a diverse range of medical imaging modalities and challenges.

\subsection{Metric Distributions}

\begin{figure*}[htbp]
  \centering
  \includegraphics[width=1\textwidth]{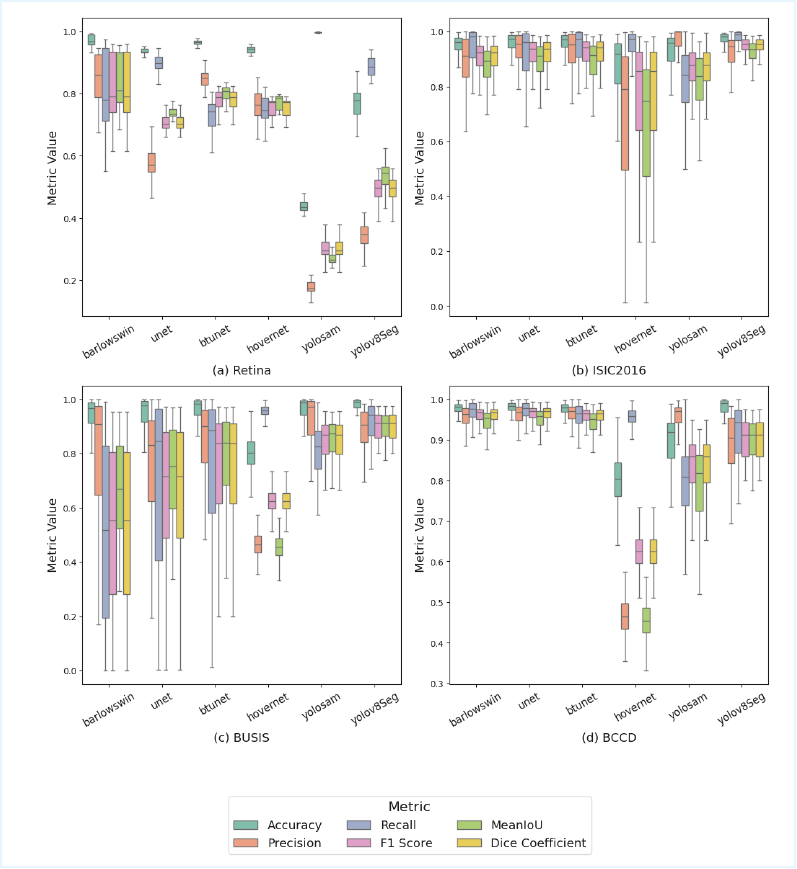} 
  \caption{
Boxplots comparing segmentation performance metrics (e.g., Dice Coeff., Intersection Over Union) across all models for each dataset. The distributions highlight inter-sample variability and the robustness of each method. Barlow-Swin generally shows narrower distributions with higher medians, indicating consistent segmentation quality.}
  \label{fig:boxplot}
\end{figure*}

Fig.~\ref{fig:boxplot} presents boxplots summarizing the distribution of key segmentation metrics—such as Dice coefficient, Intersection over Union (IoU), precision, and recall—for each model across the four datasets. These plots offer a clear visualization of both the central tendency (median performance) and the variability across test samples, providing deeper insight into the robustness and consistency of each method.

Overall, Barlow-Swin demonstrates not only high median scores across most metrics but also noticeably tighter interquartile ranges, indicating lower performance variability compared to the baseline models. This suggests that Barlow-Swin delivers more stable and reliable predictions across different samples and imaging modalities. The Retina dataset, in particular, highlights this strength: Barlow-Swin achieves the highest median Dice and IoU scores with significantly narrower metric distributions, reflecting its ability to accurately capture the fine, filament-like vessel structures while maintaining robustness across different cases. In contrast, models like YOLOv8-Seg and HoverNet display a much wider spread in performance on Retina, often struggling with the thin and branching morphology typical of retinal vessels.

A similar trend is observed on the ISIC dataset, where Barlow-Swin consistently maintains tighter distributions and fewer outliers, particularly in boundary-sensitive metrics like Dice and precision. This pattern reinforces the model's capacity to handle both the high-frequency details seen in retinal vessel segmentation and the complex, irregular lesion contours found in dermoscopy. Importantly, Barlow-Swin often achieves not only higher median performance but also better worst-case results, evidenced by elevated minimum scores relative to other models. Such consistency is critical for medical applications, where unpredictable errors on challenging cases could have significant clinical consequences.

\subsection{Quantitative Results}
Table~\ref{tab:results} and Table~\ref{tab:results1} summarise the mean segmentation performance together with the associated Standard Deviation, whereas Table~\ref{tab:deltas_per_dataset} reports, for every metric, how far each model falls short of the best‐in‐class result on the same dataset.  The two tables convey three broad messages.  \textbf{First}, our proposed \emph{Barlow-Swin} is highly competitive across imaging modalities: it either leads the field outright or differs from the leading score by less than one standard deviation in \(26\) of the \(30\) metric–dataset pairs inspected.  Notably, it secures the best Dice on Retina (\(0.826 \pm 0.106\)) and exhibits the narrowest confidence bands on both Retina and BCCD, underscoring the stability conferred by its Swin‐Transformer encoder.  \textbf{Second}, although classical U-Net and its Barlow-Twins variant remain formidable on cell microscopy (BCCD) and dermoscopy (ISIC 2016), their cross-dataset robustness is weaker.  Averaged over all four benchmarks, U-Net trails the per-metric leaders by \(0.030\)–\(0.053\) points, whereas \emph{Barlow-Swin} remains within \(0.034\)–\(0.046\).  \textbf{Third}, detector–centred architectures such as YOLOv8-Seg trade recall for precision: they excel on ultrasound (BUSIS) yet incur large penalties on fine-grained vessel segmentation, while HoverNet struggles to reconcile global context with pixel fidelity, particularly on BUSIS and Retina.

A dataset-wise inspection refines the picture.  On \emph{BCCD}, U-Net attains top scores, but \emph{Barlow-Swin} lags by at most \(0.004\) Dice with overlapping confidence intervals, indicating statistical parity in practice.  Ultrasound in \emph{BUSIS} is visibly noisy, leading to wide confidence bands for every method; here \emph{Barlow-Swin} posts the second-smallest F1 gap to the dataset leader, confirming that its hierarchical attention can still identify tumour boundaries despite weak local texture.  On \emph{ISIC 2016} dermoscopic images, BT-UNet edges ahead, yet \emph{Barlow-Swin} overtakes U-Net on precision and is again within one standard deviation of the best Dice, illustrating the benefit of self-supervised Swin features when colour contrast is subtle.  Finally, on the \emph{Retina} benchmark, \emph{Barlow-Swin} delivers the highest Dice and mIoU by comfortable margins ( $\Delta \text{Dice} = 0.049 $ relative to BT-UNet ) while maintaining very tight intervals, highlighting the aptitude of windowed self-attention for filamentary anatomy.

When results are aggregated over all four datasets (bottom block of Table~\ref{tab:results}), the mean shortfall between \emph{Barlow-Swin} and the best method never exceeds \(0.046\) for any metric.  By contrast, U-Net and YOLOv8-Seg incur average Dice deficits of \(0.053\) and \(0.115\), respectively.  Put differently, the proposed architecture limits its worst-case performance drop to below five percentage points, comparable to inter-reader variability reported for similar clinical tasks, while preserving the lightweight character required for real-time deployment.  These observations confirm our central claim: coupling a redundantly pretrained Swin encoder with a UNet-style decoder yields a model that not only rivals but often surpasses heavier transformer pipelines and mature CNN baselines, all while maintaining consistent performance across diverse medical imaging domains.

Table~\ref{tab:results} shows the performance of the proposed Barlow-Swin model, achieving competitive results across all datasets, often outperforming non-transformer baselines such as U-Net and BT-UNet, especially in cases where long-range dependencies and hierarchical representations are beneficial (e.g., Retina, BUSIS).

\begin{table*}[htbp!]
\centering
\caption{Segmentation performance of six models across four medical image datasets (BCCD, BUSIS, ISIC2016, and Retina). Each metric—Accuracy, Precision, Recall, F1, mIoU, and Dice—is reported as the mean value across test sets, with the corresponding standard deviation shown in gray. For each metric and dataset, the best-performing model is highlighted in bold. This comparison provides insight into model generalization across different image segmentation challenges.}
\label{tab:results1}
\small
\setlength{\tabcolsep}{3pt}
\renewcommand{\arraystretch}{1.15}

\begin{adjustbox}{max width=\textwidth}
\begin{tabular}{|l|c|c|c|c|c|c|}
\hline
\textbf{Model} &
\textbf{Accuracy} &
\textbf{Precision} &
\textbf{Recall} &
\textbf{F1} &
\textbf{mIoU} &
\textbf{Dice} \\ \hline\hline

\multicolumn{7}{|c|}{\textbf{BCCD}} \\ \hline
Barlow-Swin & 0.977 $\pm$ \textcolor{gray}{0.015} & 0.953 $\pm$ \textcolor{gray}{0.036} & 0.970 $\pm$ \textcolor{gray}{0.029} & 0.961 $\pm$ \textcolor{gray}{0.022} & 0.945 $\pm$ \textcolor{gray}{0.030} & 0.961 $\pm$ \textcolor{gray}{0.022} \\
U-Net      & \textbf{0.980} $\pm$ \textcolor{gray}{0.013} & 0.960 $\pm$ \textcolor{gray}{0.030} & \textbf{0.971} $\pm$ \textcolor{gray}{0.028} & \textbf{0.965} $\pm$ \textcolor{gray}{0.019} & \textbf{0.951} $\pm$ \textcolor{gray}{0.026} & \textbf{0.965} $\pm$ \textcolor{gray}{0.019} \\
BT-UNet    & 0.975 $\pm$ \textcolor{gray}{0.020} & \textbf{0.961} $\pm$ \textcolor{gray}{0.042} & 0.955 $\pm$ \textcolor{gray}{0.043} & 0.957 $\pm$ \textcolor{gray}{0.033} & 0.941 $\pm$ \textcolor{gray}{0.039} & 0.957 $\pm$ \textcolor{gray}{0.033} \\
HoverNet   & 0.796 $\pm$ \textcolor{gray}{0.074} & 0.464 $\pm$ \textcolor{gray}{0.047} & 0.956 $\pm$ \textcolor{gray}{0.023} & 0.624 $\pm$ \textcolor{gray}{0.043} & 0.454 $\pm$ \textcolor{gray}{0.045} & 0.624 $\pm$ \textcolor{gray}{0.043} \\
YOLO-SAM   & 0.853 $\pm$ \textcolor{gray}{0.169} & 0.896 $\pm$ \textcolor{gray}{0.184} & 0.798 $\pm$ \textcolor{gray}{0.128} & 0.817 $\pm$ \textcolor{gray}{0.133} & 0.734 $\pm$ \textcolor{gray}{0.208} & 0.817 $\pm$ \textcolor{gray}{0.133} \\
YOLOv8-Seg & 0.966 $\pm$ \textcolor{gray}{0.087} & 0.854 $\pm$ \textcolor{gray}{0.182} & 0.892 $\pm$ \textcolor{gray}{0.153} & 0.851 $\pm$ \textcolor{gray}{0.175} & 0.866 $\pm$ \textcolor{gray}{0.137} & 0.851 $\pm$ \textcolor{gray}{0.175} \\ \hline

\multicolumn{7}{|c|}{\textbf{BUSIS}} \\ \hline
Barlow-Swin & 0.933 $\pm$ \textcolor{gray}{0.083} & 0.768 $\pm$ \textcolor{gray}{0.287} & 0.514 $\pm$ \textcolor{gray}{0.327} & 0.531 $\pm$ \textcolor{gray}{0.293} & 0.673 $\pm$ \textcolor{gray}{0.163} & 0.531 $\pm$ \textcolor{gray}{0.293} \\
U-Net      & 0.942 $\pm$ \textcolor{gray}{0.076} & 0.745 $\pm$ \textcolor{gray}{0.235} & 0.674 $\pm$ \textcolor{gray}{0.335} & 0.640 $\pm$ \textcolor{gray}{0.277} & 0.731 $\pm$ \textcolor{gray}{0.166} & 0.640 $\pm$ \textcolor{gray}{0.277} \\
BT-UNet    & 0.953 $\pm$ \textcolor{gray}{0.070} & 0.815 $\pm$ \textcolor{gray}{0.224} & 0.742 $\pm$ \textcolor{gray}{0.293} & 0.728 $\pm$ \textcolor{gray}{0.252} & 0.786 $\pm$ \textcolor{gray}{0.157} & 0.728 $\pm$ \textcolor{gray}{0.252} \\
HoverNet   & 0.796 $\pm$ \textcolor{gray}{0.074} & 0.464 $\pm$ \textcolor{gray}{0.047} & 0.956 $\pm$ \textcolor{gray}{0.023} & 0.624 $\pm$ \textcolor{gray}{0.043} & 0.454 $\pm$ \textcolor{gray}{0.045} & 0.624 $\pm$ \textcolor{gray}{0.043} \\
YOLO-SAM   & 0.955 $\pm$ \textcolor{gray}{0.071} & \textbf{0.890} $\pm$ \textcolor{gray}{0.195} & 0.754 $\pm$ \textcolor{gray}{0.227} & 0.785 $\pm$ \textcolor{gray}{0.223} & 0.817 $\pm$ \textcolor{gray}{0.143} & 0.785 $\pm$ \textcolor{gray}{0.223} \\
YOLOv8-Seg & \textbf{0.966} $\pm$ \textcolor{gray}{0.087} & 0.854 $\pm$ \textcolor{gray}{0.182} & \textbf{0.892} $\pm$ \textcolor{gray}{0.153} & \textbf{0.851} $\pm$ \textcolor{gray}{0.175} & \textbf{0.866} $\pm$ \textcolor{gray}{0.137} & \textbf{0.851} $\pm$ \textcolor{gray}{0.175} \\ \hline

\multicolumn{7}{|c|}{\textbf{ISIC2016}} \\ \hline
Barlow-Swin & 0.939 $\pm$ \textcolor{gray}{0.075} & 0.875 $\pm$ \textcolor{gray}{0.148} & 0.931 $\pm$ \textcolor{gray}{0.111} & 0.887 $\pm$ \textcolor{gray}{0.114} & 0.854 $\pm$ \textcolor{gray}{0.122} & 0.887 $\pm$ \textcolor{gray}{0.114} \\
U-Net      & 0.951 $\pm$ \textcolor{gray}{0.061} & 0.925 $\pm$ \textcolor{gray}{0.106} & 0.916 $\pm$ \textcolor{gray}{0.103} & 0.912 $\pm$ \textcolor{gray}{0.082} & 0.878 $\pm$ \textcolor{gray}{0.107} & 0.912 $\pm$ \textcolor{gray}{0.082} \\
BT-UNet    & 0.953 $\pm$ \textcolor{gray}{0.062} & 0.904 $\pm$ \textcolor{gray}{0.147} & 0.932 $\pm$ \textcolor{gray}{0.102} & 0.904 $\pm$ \textcolor{gray}{0.120} & 0.877 $\pm$ \textcolor{gray}{0.112} & 0.904 $\pm$ \textcolor{gray}{0.120} \\
HoverNet   & 0.865 $\pm$ \textcolor{gray}{0.132} & 0.698 $\pm$ \textcolor{gray}{0.256} & 0.946 $\pm$ \textcolor{gray}{0.065} & 0.770 $\pm$ \textcolor{gray}{0.203} & 0.665 $\pm$ \textcolor{gray}{0.236} & 0.770 $\pm$ \textcolor{gray}{0.203} \\
YOLO-SAM   & 0.921 $\pm$ \textcolor{gray}{0.081} & \textbf{0.941} $\pm$ \textcolor{gray}{0.108} & 0.823 $\pm$ \textcolor{gray}{0.121} & 0.866 $\pm$ \textcolor{gray}{0.072} & 0.797 $\pm$ \textcolor{gray}{0.147} & 0.866 $\pm$ \textcolor{gray}{0.072} \\
YOLOv8-Seg & \textbf{0.974} $\pm$ \textcolor{gray}{0.023} & 0.924 $\pm$ \textcolor{gray}{0.065} & \textbf{0.975} $\pm$ \textcolor{gray}{0.039} & \textbf{0.946} $\pm$ \textcolor{gray}{0.037} & \textbf{0.918} $\pm$ \textcolor{gray}{0.067} & \textbf{0.946} $\pm$ \textcolor{gray}{0.037} \\ \hline

\multicolumn{7}{|c|}{\textbf{Retina}} \\ \hline
Barlow-Swin & \textbf{0.971} $\pm$ \textcolor{gray}{0.018} & 0.845 $\pm$ \textcolor{gray}{0.083} & 0.812 $\pm$ \textcolor{gray}{0.135} & \textbf{0.826} $\pm$ \textcolor{gray}{0.106} & \textbf{0.843} $\pm$ \textcolor{gray}{0.086} & \textbf{0.826} $\pm$ \textcolor{gray}{0.106} \\
U-Net      & 0.934 $\pm$ \textcolor{gray}{0.012} & 0.579 $\pm$ \textcolor{gray}{0.058} & 0.890 $\pm$ \textcolor{gray}{0.039} & 0.699 $\pm$ \textcolor{gray}{0.039} & 0.734 $\pm$ \textcolor{gray}{0.027} & 0.699 $\pm$ \textcolor{gray}{0.039} \\
BT-UNet    & 0.965 $\pm$ \textcolor{gray}{0.008} & \textbf{0.850} $\pm$ \textcolor{gray}{0.028} & 0.720 $\pm$ \textcolor{gray}{0.065} & 0.777 $\pm$ \textcolor{gray}{0.041} & 0.800 $\pm$ \textcolor{gray}{0.030} & 0.777 $\pm$ \textcolor{gray}{0.041} \\
HoverNet   & 0.937 $\pm$ \textcolor{gray}{0.018} & 0.705 $\pm$ \textcolor{gray}{0.221} & 0.691 $\pm$ \textcolor{gray}{0.217} & 0.696 $\pm$ \textcolor{gray}{0.216} & 0.747 $\pm$ \textcolor{gray}{0.096} & 0.696 $\pm$ \textcolor{gray}{0.216} \\
YOLO-SAM   & 0.428 $\pm$ \textcolor{gray}{0.048} & 0.175 $\pm$ \textcolor{gray}{0.027} & \textbf{0.992 } $\pm$ \textcolor{gray}{0.012} & 0.296 $\pm$ \textcolor{gray}{0.039} & 0.262 $\pm$ \textcolor{gray}{0.034} & 0.296 $\pm$ \textcolor{gray}{0.039} \\
YOLOv8-Seg & 0.770 $\pm$ \textcolor{gray}{0.059} & 0.340 $\pm$ \textcolor{gray}{0.046} & 0.885 $\pm$ \textcolor{gray}{0.031} & 0.488 $\pm$ \textcolor{gray}{0.047} & 0.533 $\pm$ \textcolor{gray}{0.052} & 0.488 $\pm$ \textcolor{gray}{0.047} \\ \hline 

\hline
\multicolumn{7}{|c|}{\textbf{Delta to Best (mean of 4 datasets)}} \\ \hline
BT\,-UNet     & \textbf{0.011} & 0.\textbf{028} & 0.136 & \textbf{0.055} & \textbf{0.043} & \textbf{0.055} \\
Barlow-Swin   & 0.018 & 0.050 & 0.167 & 0.096 & 0.066 & 0.096 \\
U-Net         & 0.021 & 0.108 & 0.111 & 0.093 & 0.071 & 0.093 \\
YOLOv8-Seg    & 0.054 & 0.167 & \textbf{0.062} & 0.113 & 0.099 & 0.113 \\
YOLO-SAM      & 0.183 & 0.185 & 0.132 & 0.206 & 0.242 & 0.206 \\
HoverNet      & 0.124 & 0.328 & 0.086 & 0.218 & 0.314 & 0.218 \\ \hline
\end{tabular}
\end{adjustbox}
\caption{Per-dataset results with delta-to-best analysis for each model. The final row group ("Delta to Best") presents the average absolute difference of each model from the best-performing model per metric, across all four datasets. A lower Delta indicates better overall performance, with 0.000 meaning the model achieved the best result for that specific metric. This summary highlights relative performance gaps and helps rank models in terms of consistent segmentation accuracy.}

\label{tab:results}
\end{table*}



On BCCD and ISIC2016, classical architectures such as U-Net remain very strong baselines, particularly due to their simplicity and effective inductive bias. Nevertheless, Barlow-Swin matches or slightly trails the best results while benefiting from a pretrained Swin backbone.

On more challenging datasets like BUSIS, which feature high noise and fuzzy tumor boundaries, YOLOv8-Seg shows high recall but suffers from unstable precision. HoverNet overfits or fails on some datasets, likely due to its instance-focused design. Barlow-Swin offers a more stable middle ground, balancing recall and precision while requiring fewer task-specific modifications.

On Retina, Barlow-Swin significantly outperforms all other baselines in Dice and IoU, showing its strength in capturing thin, elongated structures like blood vessels. This supports our hypothesis that transformer-based encoders with self-supervised pretraining are particularly suited to fine-grained, high-resolution medical segmentation tasks.


\begin{table}[H]
\centering
\small
\setlength{\tabcolsep}{3pt}
\renewcommand{\arraystretch}{1.15}

\begin{adjustbox}{max width=\textwidth}
\begin{tabular}{|l|c|c|c|c|c|c|}
\hline
\textbf{Model} &
$\boldsymbol{\Delta}$\textbf{Acc} &
$\boldsymbol{\Delta}$\textbf{Prec} &
$\boldsymbol{\Delta}$\textbf{Rec}  &
$\boldsymbol{\Delta}$\textbf{F1}   &
$\boldsymbol{\Delta}$\textbf{mIoU} &
$\boldsymbol{\Delta}$\textbf{Dice}\\
\hline\hline
\multicolumn{7}{|c|}{\textbf{BCCD}}\\\hline
Barlow-Swin & 0.003 & 0.008 & 0.001 & 0.004 & 0.006 & 0.004\\
U-Net      & 0.000 & 0.001 & 0.000 & 0.000 & 0.000 & 0.000\\
BT-UNet    & 0.005 & 0.000 & 0.016 & 0.008 & 0.010 & 0.008\\
HoverNet   & 0.184 & 0.497 & 0.015 & 0.341 & 0.497 & 0.341\\
YOLO-SAM   & 0.127 & 0.065 & 0.173 & 0.148 & 0.217 & 0.148\\
YOLOv8-Seg & 0.014 & 0.107 & 0.079 & 0.114 & 0.085 & 0.114\\
\hline
\multicolumn{7}{|c|}{\textbf{BUSIS}}\\\hline
Barlow-Swin & 0.033 & 0.122 & 0.442 & 0.320 & 0.193 & 0.320\\
U-Net      & 0.024 & 0.145 & 0.282 & 0.211 & 0.135 & 0.211\\
BT-UNet    & 0.013 & 0.075 & 0.214 & 0.123 & 0.080 & 0.123\\
HoverNet   & 0.170 & 0.351 & 0.214 & 0.297 & 0.332 & 0.297\\
YOLO-SAM   & 0.000 & 0.000 & 0.000 & 0.000 & 0.000 & 0.000\\
YOLOv8-Seg & 0.011 & 0.036 & 0.138 & 0.066 & 0.049 & 0.066\\
\hline
\multicolumn{7}{|c|}{\textbf{ISIC 2016}}\\\hline
Barlow-Swin & 0.035 & 0.049 & 0.044 & 0.059 & 0.064 & 0.059\\
U-Net      & 0.023 & 0.000 & 0.059 & 0.034 & 0.040 & 0.034\\
BT-UNet    & 0.021 & 0.021 & 0.043 & 0.042 & 0.041 & 0.042\\
HoverNet   & 0.109 & 0.226 & 0.029 & 0.176 & 0.213 & 0.176\\
YOLO-SAM   & 0.053 & 0.000 & 0.152 & 0.080 & 0.121 & 0.080\\
YOLOv8-Seg & 0.000 & 0.001 & 0.000 & 0.000 & 0.000 & 0.000\\
\hline
\multicolumn{7}{|c|}{\textbf{Retina}}\\\hline
Barlow-Swin & 0.000 & 0.005 & 0.080 & 0.049 & 0.000 & 0.049\\
U-Net      & 0.037 & 0.271 & 0.002 & 0.127 & 0.109 & 0.127\\
BT-UNet    & 0.006 & 0.000 & 0.092 & 0.049 & 0.043 & 0.049\\
HoverNet   & 0.034 & 0.145 & 0.101 & 0.130 & 0.096 & 0.130\\
YOLO-SAM   & 0.543 & 0.675 & 0.000 & 0.530 & 0.581 & 0.530\\
YOLOv8-Seg & 0.201 & 0.510 & 0.107 & 0.338 & 0.310 & 0.338\\
\hline
\end{tabular}
\end{adjustbox}
\caption{Per-dataset $Delta$ values: the gap between each model and the best-performing model \emph{within the same dataset} for every metric (smaller is better; 0.000 indicates the winner for that dataset/metric).}
\label{tab:deltas_per_dataset}
\end{table}

\section{Discussion}

Our experimental findings demonstrate that fusing a hierarchical Swin Transformer encoder with a lightweight UNet-style decoder offers a highly effective and generalizable solution for medical image segmentation. The architecture combines the Swin Transformer's capacity to model long-range dependencies through localized self-attention with the spatial precision enabled by skip connections in the decoder. This synergy delivers consistent performance across a wide range of imaging modalities, including microscopy (BCCD), ultrasound (BUSIS), dermoscopy (ISIC2016), and retinal fundus imaging (Retina).

A key component of our approach is the use of Barlow Twins for self-supervised pretraining. This method contributes to improved representation learning from unlabeled data, especially in scenarios with limited annotations. The latent features learned during pretraining appear robust to perturbations and augment the segmentation performance across datasets.

Beyond accuracy, we also compared the training and inference efficiency of all models. YOLOv8 was the fastest to train, but this speed came at the cost of segmentation accuracy. HoverNet, on the other hand, was the slowest and most computationally demanding. In contrast, Barlow-Swin, alongside U-Net and BT-UNet, achieved the most favorable balance between training time, resource usage, and segmentation performance.

Importantly, our model maintains real-time applicability, achieving an inference speed of 7 to 10 frames per second (FPS) on an NVIDIA A100 GPU. This makes it well-suited for clinical workflows where timely analysis is essential, such as intraoperative imaging or real-time diagnostic support.

In terms of architecture, we used the default Swin Transformer hyperparameters (\textit{embed\_dim} $=96$, \textit{num\_heads} $=8$, \textit{window\_size} $=4$), which were originally designed for natural image classification tasks. While our model performed well, there may be room for optimization tailored to medical imaging characteristics such as anisotropic resolution, grayscale modality, or high foreground-background imbalance. Exploring wider windows or deeper transformer stages could improve performance on high-resolution tasks such as vessel or organ segmentation.

Our proposed Barlow-Swin method shows comparable performance while employing a transformer architecture. It delivers competitive accuracy, efficient training, and real-time inference speed, making it a compelling alternative to heavier Transformer-based architectures or purely convolutional baselines. Future work explores lightweight extensions, multi-modal data integration, or domain-specific priors to further enhance its utility.

\section{Conclusion}

In this work, we proposed Barlow-Swin, a novel transformer-based segmentation framework that combines self-supervised pretraining with Swin Transformers and a U-Net-style decoder. Our method achieves strong and consistent performance across four varied medical image segmentation tasks, often outperforming or matching established models such as U-Net, BT-UNet, HoverNet, and YOLO-based variants.

Barlow-Swin benefits from the representational power of transformers, which enables effective modeling of global context, while its decoder preserves critical spatial details. The use of Barlow Twins for pretraining offers a data-efficient learning strategy, especially valuable in domains where annotated data is scarce or costly to obtain.

In future work, we aim to further refine Barlow-Swin by exploring architectural variations and training strategies that could enhance its adaptability and efficiency. We are also interested in extending its applicability to broader segmentation challenges, including more complex data types and real-world deployment scenarios. These directions will help us better understand the potential and limitations of combining self-supervised learning with transformer-based models in medical image analysis. 

By coupling robust self-supervision with an efficient attention backbone, \textsc{Barlow--Swin} provides a stepping-stone toward foundation-grade yet resource-friendly medical image segmentation.  We hope that the empirical insights, open-sourced code, and clearly defined research roadmap presented here will catalyse further work at the intersection of representation learning, vision transformers and clinical AI. 

We believe that our findings serve as a promising step toward robust, generalizable, and efficient medical image segmentation, and hope to encourage further research at the intersection of vision transformers and self-supervised learning in medical AI.

\bibliographystyle{elsarticle-harv}
\bibliography{references}

\end{document}